# Automating psychological hypothesis generation with AI: when large language models meet causal graph

Song Tong[1,2,3,4,6], Kai Mao[5,6], Zhen Huang[2], Yukun Zhao[2]✉ & Kaiping Peng[1,2,3,4]✉

Leveraging the synergy between causal knowledge graphs and a large language model (LLM), our study introduces a groundbreaking approach for computational hypothesis generation in psychology. We analyzed 43,312 psychology articles using a LLM to extract causal relation pairs. This analysis produced a specialized causal graph for psychology. Applying link prediction algorithms, we generated 130 potential psychological hypotheses focusing on "wellbeing", then compared them against research ideas conceived by doctoral scholars and those produced solely by the LLM. Interestingly, our combined approach of a LLM and causal graphs mirrored the expert-level insights in terms of novelty, clearly surpassing the LLM-only hypotheses ($t(59) = 3.34$, $p = 0.007$ and $t(59) = 4.32$, $p < 0.001$, respectively). This alignment was further corroborated using deep semantic analysis. Our results show that combining LLM with machine learning techniques such as causal knowledge graphs can revolutionize automated discovery in psychology, extracting novel insights from the extensive literature. This work stands at the crossroads of psychology and artificial intelligence, championing a new enriched paradigm for data-driven hypothesis generation in psychological research.

[1] Department of Psychological and Cognitive Sciences, Tsinghua University, Beijing, China. [2] Positive Psychology Research Center, School of Social Sciences, Tsinghua University, Beijing, China. [3] AI for Wellbeing Lab, Tsinghua University, Beijing, China. [4] Institute for Global Industry, Tsinghua University, Beijing, China. [5] Kindom KK, Tokyo, Japan. [6] These authors contributed equally: Song Tong, Kai Mao. ✉email: zhaoyukun@tsinghua.edu.cn; pengkp@tsinghua.edu.cn

## Introduction

In an age in which the confluence of artificial intelligence (AI) with various subjects profoundly shapes sectors ranging from academic research to commercial enterprises, dissecting the interplay of these disciplines becomes paramount (Williams et al., 2023). In particular, psychology, which serves as a nexus between the humanities and natural sciences, consistently endeavors to demystify the complex web of human behaviors and cognition (Hergenhahn and Henley, 2013). Its profound insights have significantly enriched academia, inspiring innovative applications in AI design. For example, AI models have been molded on hierarchical brain structures (Cichy et al., 2016) and human attention systems (Vaswani et al., 2017). Additionally, these AI models reciprocally offer a rejuvenated perspective, deepening our understanding from the foundational cognitive taxonomy to nuanced esthetic perceptions (Battleday et al., 2020; Tong et al., 2021). Nevertheless, the multifaceted domain of psychology, particularly social psychology, has exhibited a measured evolution compared to its tech-centric counterparts. This can be attributed to its enduring reliance on conventional theory-driven methodologies (Henrich et al., 2010; Shah et al., 2015), a characteristic that stands in stark contrast to the burgeoning paradigms of AI and data-centric research (Bechmann and Bowker, 2019; Wang et al., 2023).

In the journey of psychological research, each exploration originates from a spark of innovative thought. These research trajectories may arise from established theoretical frameworks, daily event insights, anomalies within data, or intersections of interdisciplinary discoveries (Jaccard and Jacoby, 2019). Hypothesis generation is pivotal in psychology (Koehler, 1994; McGuire, 1973), as it facilitates the exploration of multifaceted influencers of human attitudes, actions, and beliefs. The HyGene model (Thomas et al., 2008) elucidated the intricacies of hypothesis generation, encompassing the constraints of working memory and the interplay between ambient and semantic memories. Recently, causal graphs have provided psychology with a systematic framework that enables researchers to construct and simulate intricate systems for a holistic view of "bio-psycho-social" interactions (Borsboom et al., 2021; Crielaard et al., 2022). Yet, the labor-intensive nature of the methodology poses challenges, which requires multidisciplinary expertise in algorithmic development, exacerbating the complexities (Crielaard et al., 2022). Meanwhile, advancements in AI, exemplified by models such as the generative pretrained transformer (GPT), present new avenues for creativity and hypothesis generation (Wang et al., 2023).

Building on this, notably large language models (LLMs) such as GPT-3, GPT-4, and Claude-2, which demonstrate profound capabilities to comprehend and infer causality from natural language texts, a promising path has emerged to extract causal knowledge from vast textual data (Binz and Schulz, 2023; Gu et al., 2023). Exciting possibilities are seen in specific scenarios in which LLMs and causal graphs manifest complementary strengths (Pan et al., 2023). Their synergistic combination converges human analytical and systemic thinking, echoing the holistic versus analytic cognition delineated in social psychology (Nisbett et al., 2001). This amalgamation enables fine-grained semantic analysis and conceptual understanding via LLMs, while causal graphs offer a global perspective on causality, alleviating the interpretability challenges of AI (Pan et al., 2023). This integrated methodology efficiently counters the inherent limitations of working and semantic memories in hypothesis generation and, as previous academic endeavors indicate, has proven efficacious across disciplines. For example, a groundbreaking study in physics synthesized 750,000 physics publications, utilizing cutting-edge natural language processing to extract 6368 pivotal quantum physics concepts, culminating in a semantic network forecasting research trajectories (Krenn and Zeilinger, 2020). Additionally, by integrating knowledge-based causal graphs into the foundation of the LLM, the LLM's capability for causative inference significantly improves (Kıcıman et al., 2023).

To this end, our study seeks to build a pioneering analytical framework, combining the semantic and conceptual extraction proficiency of LLMs with the systemic thinking of the causal graph, with the aim of crafting a comprehensive causal network of semantic concepts within psychology. We meticulously analyzed 43,312 psychological articles, devising an automated method to construct a causal graph, and systematically mining causative concepts and their interconnections. Specifically, the initial sifting and preparation of the data ensures a high-quality corpus, and is followed by employing advanced extraction techniques to identify standardized causal concepts. This results in a graph database that serves as a reservoir of causal knowledge. In conclusion, using node embedding and similarity-based link prediction, we unearthed potential causal relationships, and thus generated the corresponding hypotheses.

To gauge the pragmatic value of our network, we selected 130 hypotheses on "well-being" generated by our framework, comparing them with hypotheses crafted by novice experts (doctoral students in psychology) and the LLM models. The results are encouraging: Our algorithm matches the caliber of novice experts, outshining the hypotheses generated solely by the LLM models in novelty. Additionally, through deep semantic analysis, we demonstrated that our algorithm contains more profound conceptual incorporations and a broader semantic spectrum.

Our study advances the field of psychology in two significant ways. Firstly, it extracts invaluable causal knowledge from the literature and converts it to visual graphics. These aids can feed algorithms to help deduce more latent causal relations and guide models in generating a plethora of novel causal hypotheses. Secondly, our study furnishes novel tools and methodologies for causal analysis and scientific knowledge discovery, representing the seamless fusion of modern AI with traditional research methodologies. This integration serves as a bridge between conventional theory-driven methodologies in psychology and the emerging paradigms of data-centric research, thereby enriching our understanding of the factors influencing psychology, especially within the realm of social psychology.

## Methodological framework for hypothesis generation

The proposed LLM-based causal graph (LLMCG) framework encompasses three steps: literature retrieval, causal pair extraction, and hypothesis generation, as illustrated in Fig. 1. In the literature gathering phase, ~140k psychology-related articles were downloaded from public databases. In step two, GPT-4 were used to distil causal relationships from these articles, culminating in the creation of a causal relationship network based on 43,312 selected articles. In the third step, an in-depth examination of these data was executed, adopting link prediction algorithms to forecast the dynamics within the causal relationship network for searching the highly potential causality concept pairs.

**Step 1: Literature retrieval.** The primary data source for this study was a public repository of scientific articles, the PMC Open Access Subset. Our decision to utilize this repository was informed by several key attributes that it possesses. The PMC Open Access Subset boasts an expansive collection of over 2 million full-text XML science and medical articles, providing a substantial and diverse base from which to derive insights for our research. Furthermore, the open-access nature of the articles not only enhances the transparency and reproducibility of our methodology, but also ensures that the results and processes can be independently accessed and verified by other researchers. Notably, the content within this subset originates from recognized journals, all of which have undergone rigorous peer review, lending credence to the quality and reliability of the data we leveraged. Finally, an added advantage was the rich metadata accompanying each article. These metadata were instrumental in refining our article selection process, ensuring coherent thematic alignment with our research objectives in the domains of psychology.

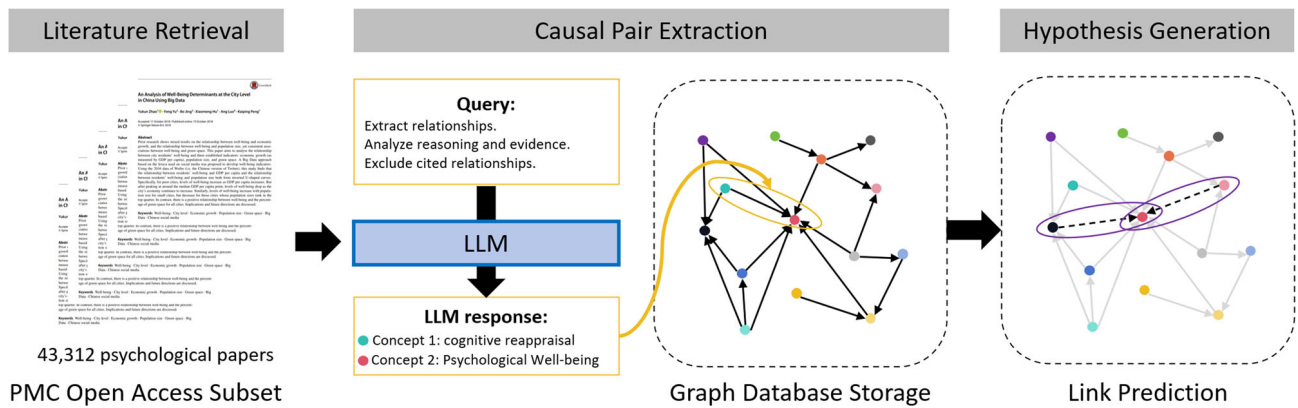

**Fig. 1 The hypothesis generation framework Using LLMCG Algorithm.** Note: LLM stands for large language model; LLMCG algorithm stands for LLM-based causal graph algorithm, which includes the processes of literature retrieval, causal pair extraction, and hypothesis generation.

To identify articles relevant to our study, we applied a series of filtering criteria. First, the presence of certain keywords within article titles or abstracts was mandatory. Some examples of these keywords include "psychol", "clin psychol", and "biol psychol". Second, we exploited the metadata accompanying each article. The classification of articles based on these metadata ensured alignment with recognized thematic standards in the domains of psychology and neuroscience. Upon the application of these criteria, we managed to curate a subset of approximately 140K articles that most likely discuss causal concepts in both psychology and neuroscience.

**Step 2: Causal pair extraction.** The process of extracting causal knowledge from vast troves of scientific literature is intricate and multifaceted. Our methodology distils this complex process into four coherent steps, each serving a distinct purpose. (1) Article selection and cost analysis: Determines the feasibility of processing a specific volume of articles, ensuring optimal resource allocation. (2) Text extraction and analysis: Ensures the purity of the data that enter our causal extraction phase by filtering out nonrelevant content. (3) Causal knowledge extraction: Uses advanced language models to detect, classify, and standardize causal factors relationships present in texts. (4) Graph database storage: Facilitates structured storage, easy retrieval, and the possibility of advanced relational analyses for future research. This streamlined approach ensures accuracy, consistency, and scalability in our endeavor to understand the interplay of causal concepts in psychology and neuroscience.

*Text extraction and cleaning.* After a meticulous cost analysis detailed in Appendix A, our selection process identified 43,312 articles. This selection was strategically based on the criterion that the journal titles must incorporate the term "Psychol", signifying their direct relevance to the field of psychology. The distributions of publication sources and years can be found in Table 1. Extracting the full texts of the articles from their PDF sources was

| Table 1 Data sources and publication year distribution for LLMCG computation. | |
|---|---:|
| | Quantity |
| **Source of Publications** | |
| Frontiers in Psychology | 35,797 |
| BMC Psychology | 1154 |
| Journal of Psychological Medicine and Mental Pathology | 687 |
| Psychological Medicine | 659 |
| European Journal of Investigation in Health Psychology and Education | 384 |
| European Journal of Psychology | 369 |
| Psychological Research | 305 |
| Health Psychology and Behavioral Medicine | 277 |
| Current Psychology | 227 |
| Journal of Child Psychology and Psychiatry | 202 |
| Others | 3251 |
| Total | 43,312 |
| **Publication Year (1975–2023)** | |
| 2023 | 2391 |
| 2022 | 9876 |
| 2021 | 7286 |
| 2020 | 4416 |
| 2019 | 3465 |
| Others | 15,878 |
| Total | 43,312 |

an essential initial step, and, for this purpose, the PyPDF2 Python library was used. This library allowed us to seamlessly extract and concatenate titles, abstracts, and main content from each PDF article. However, a challenge arose with the presence of extraneous sections such as references or tables, in the extracted texts. The implemented procedure, employing regular expressions in Python, was not only adept at identifying variations of the term "references" but also ascertained whether this section appeared as an isolated segment. This check was critical to ensure that the identified that the "references" section was indeed distinct, marking the start of a reference list without continuation into other text. Once identified as a standalone entity, the next step in the method was to efficiently remove the reference section and its subsequent content.

*Causal knowledge extraction method.* In our effort to extract causal knowledge, the choice of GPT-4 was not arbitrary. While several models were available for such tasks, GPT-4 emerged as a frontrunner due to its advanced capabilities (Wu et al., 2023), extensive training on diverse data, with its proven proficiency in understanding context, especially in complex scientific texts

**Table 2 An example prompt and response.**

*Prompt:*
From the "text" below, extract the key causal and correlational relationships described directly in the given text by analyzing reasoning and evidence within the text. Exclude any relationships that are attributed to or cited from other research studies.
Format the relationships in JSON format with the following fields:
'concept_pair': A list representation of the cause and effect concepts in the relationship, in [cause, effect] order.
'relationship': 'causality' or 'correlation' indicating the type of relationship.
'positive/negative': If the extracted relationship is causality, indicate whether it's a positive or negative causality relationship. If it's a correlation relationship, reply as None.
{
 "PMC8451848",
 'concept_pair': [openness to change values, well-being],
 'relationship':"causality",
 'positive/negative':"positive"
},
{
 "PMC6085571",
 'concept_pair': [cognitive reappraisal, Psychological well-being],
 'relationship':"causality",
 'positive/negative':"positive"
}

(Cheng et al., 2023; Sanderson, 2023). Other models were indeed considered; however, the capacity of GPT-4 to generate coherent, contextually relevant responses gave our project an edge in its specific requirements.

The extraction process commenced with the segmentation of the articles. Due to the token constraints inherent to GPT-4, it was imperative to break down the articles into manageable chunks, specifically those of 4000 tokens or fewer. This approach ensured a comprehensive interpretation of the content without omitting any potential causal relationships. The next phase was prompt engineering. To effectively guide the extraction capabilities of GPT-4, we crafted explicit prompts. A testament to this meticulous engineering is demonstrated in a directive in which we asked the model to elucidate causal pairs in a predetermined JSON format. For a clearer understanding, readers are referred to Table 2, which elucidates the example prompt and the subsequent model response. After extraction, the outputs were not immediately cataloged. A filtering process was initiated to ascertain the standardization of the concept pairs. This process weeded out suboptimal outputs. Aiding in this quality control, GPT-4 played a pivotal role in the verification of causal pairs, determining their relevance, causality, and ensuring correct directionality. Finally, while extracting knowledge, we were aware of the constraints imposed by the GPT-4 API. There was a conscious effort to ensure that we operated within the bounds of 60 requests and 150k tokens per minute. This interplay of prompt engineering and stringent filtering was productive.

In addition, we conducted an exploratory study to assess GPT-4's discernment between "causality" and "correlation" involved four graduate students (mean age 31 ± 10.23), each evaluating relationship pairs extracted from their familiar psychology articles. The experimental details and results can be found in Appendix A and Table A1. The results showed that out of 289 relationships identified by GPT-4, 87.54% were validated. Notably, when GPT-4 classified relationships as causal, only 13.02% (31/238) were recognized as non-relationship, while 65.55% (156/238) agreed upon as causality. This shows that GPT-4 can accurately extract relationships (causality or correlation) in psychological texts, underscoring the potential as a tool for the construction of causal graphs.

To enhance the robustness of the extracted causal relationships and minimize biases, we adopted a multifaceted approach. Recognizing the indispensable role of human judgment, we periodically subjected random samples of extracted causal relationships to the scrutiny of domain experts. Their valuable feedback was instrumental in the real-time fine-tuning the extraction process. Instead of heavily relying on referenced hypotheses, our focus was on extracting causal pairs, primarily from the findings mentioned in the main texts. This systematic methodology ultimately resulted in a refined text corpus distilled from 43,312 articles, which contained many conceptual insights and were primed for rigorous causal extraction.

*Graph database storage.* Our decision to employ Neo4j as the database system was strategic. Neo4j, as a graph database (Thomer and Wickett, 2020), is inherently designed to capture and represent complex relationships between data points, an attribute that is essential for understanding intricate causal relationships. Beyond its technical prowess, Neo4j provides advantages such as scalability, resilience, and efficient querying capabilities (Webber, 2012). It is particularly adept at traversing interconnected data points, making it an excellent fit for our causal relationship analysis. The mined causal knowledge finds its abode in the Neo4j graph database. Each pair of causal concepts is represented as a node, with its directionality and interpretations stored as attributes. Relationships provide related concepts together. Storing the knowledge graph in Neo4j allows for the execution of the graph algorithms to analyze concept interconnectivity and reveal potential relationships.

The graph database contains 197k concepts and 235k connections. Table 3 encapsulates the core concepts and provides a vivid snapshot of the most recurring themes; helping us to understand the central topics that dominate the current psychological discourse. A comprehensive examination of the core concepts extracted from 43,312 psychological papers, several distinct patterns and focal areas emerged. In particular, there is a clear balance between health and illness in psychological research. The prominence of terms such as "depression", "anxiety", and "symptoms of depression magnifies the commitment in the discipline to understanding and addressing mental illnesses. However, juxtaposed against these are positive terms such as "life satisfaction" and "sense of happiness", suggesting that psychology not only fixates on challenges but also delves deeply into the nuances of positivity and well-being. Furthermore, the significance given to concepts such as "life satisfaction", "sense of happiness", and "job satisfaction" underscores an increasing recognition of emotional well-being and job satisfaction as integral to overall mental health. Intertwining the realms of psychology and neuroscience, terms such as "microglial cell activation", "cognitive impairment", and "neurodegenerative changes" signal a growing interest in understanding the neural underpinnings of cognitive and psychological phenomena. In addition, the emphasis on "self-efficacy", "positive emotions", and "self-esteem" reflect the profound interest in understanding how self-perception and emotions influence human behavior and well-being. Concepts such as "age", "resilience", and "creativity" further expand the canvas, showcasing the eclectic and comprehensive nature of inquiries in the field of psychology.

Overall, this analysis paints a vivid picture of modern psychological research, illuminating its multidimensional approach. It demonstrates a discipline that is deeply engaged with both the challenges and triumphs of human existence, offering holistic insight into the human mind and its myriad complexities.

**Step 3: Hypothesis generation using link prediction.** In the quest to uncover novel causal relationships beyond direct extraction from texts, the technique of link prediction emerges as a pivotal methodology. It hinges on the premise of proposing potential causal ties between concepts that our knowledge graph does not explicitly connect. The process intricately weaves together vector embedding, similarity analysis, and probability-based ranking. Initially, concepts are transposed into a vector space using node2vec, which is valued for its ability to capture topological nuances. Here, every pair of unconnected concepts is assigned a similarity score, and pairs that do not meet a set benchmark are quickly discarded. As we dive deeper into the higher echelons of these scored pairs, the likelihood of their linkage is assessed using the Jaccard similarity of their neighboring concepts. Subsequently, these potential causal relationships are organized in descending order of their derived probabilities, and the elite pairs are selected.

An illustration of this approach is provided in the case highlighted in Figure A1. For instance, the behavioral inhibition system (BIS) exhibits ties to both the behavioral activation system (BAS) and the subsequent behavioral response of the BAS when encountering reward stimuli, termed the BAS reward response. Simultaneously, another concept, interference, finds itself bound to both the BAS and the BAS Reward Response. This configuration hints at a plausible link between the BIS and interference. Such highly probable causal pairs are not mere intellectual curiosity. They act as springboards, catalyzing the genesis of new experimental designs or research hypotheses ripe for empirical probing. In essence, this capability equips researchers with a cutting-edge instrument, empowering them to navigate the unexplored waters of the psychological and neurological domains.

Using pairs of highly probable causal concepts, we pushed GPT-4 to conjure novel causal hypotheses that bridge concepts. To further elucidate the process of this method, Table 4 provides some examples of hypotheses generated from the process. Such hypotheses, as exemplified in the last row, underscore the potential and power of our method for generating innovative causal propositions.

### Hypotheses evaluation and results

In this section, we present an analysis focusing on quality in terms of novelty and usefulness of the hypotheses generated. According to existing literature, these dimensions are instrumental in encapsulating the essence of inventive ideas (Boden, 2009; McCarthy et al., 2018; Miron-Spektor and Beenen, 2015). These parameters have not only been quintessential for gauging creative concepts, but they have also been adopted to evaluate the caliber of research hypotheses (Dowling and Lucey, 2023; Krenn and Zeilinger, 2020; Oleinik, 2019). Specifically, we evaluate the quality of the hypotheses generated by the proposed LLMCG algorithm in relation to those generated by PhD students from an elite university who represent human junior experts, the LLM model, which represents advanced AI systems, and the research ideas refined by psychological researchers which represents cooperation between AI and humans.

**Table 3 Core concepts in 43,312 psychological papers.**

| Number | Concepts | Degree (in) |
|---|---|---|
| 1 | Depression | 2002 |
| 2 | Anxiety | 1606 |
| 3 | Life satisfaction | 890 |
| 4 | Well-being | 874 |
| 5 | Performance | 834 |
| 6 | Depressive symptoms | 812 |
| 7 | Mental health | 764 |
| 8 | Microglial activation | 734 |
| 9 | Accuracy | 720 |
| 10 | Psychological distress | 631 |
| 11 | Job satisfaction | 623 |
| 12 | Cognitive impairment | 603 |
| 13 | Neurodegeneration | 597 |
| 14 | Stress | 557 |
| 15 | Self-efficacy | 549 |
| 16 | Neuroinflammation | 541 |
| 17 | Oxidative stress | 536 |
| 18 | Age | 533 |
| 19 | Neuroprotection | 505 |
| 20 | Resilience | 492 |

**Table 4 Example hypotheses for causal relationships.**

| Concept 1 | Concept 2 | Hypothesis |
|---|---|---|
| Microbiome diversity | Well-being | Pandemic flourishing: Some individuals may experience a sense of "flourishing" or thriving during pandemic events despite the widespread stress and adversity. |
| Divergent thinking exercises | Well-being | Divergent thinking exercises can expand one's sense of self and purpose, which then synergistically improve well-being through an upward spiral. Engaging in creative thinking exercises can broaden one's perspectives on identity and meaning in life. This expanded sense of self and purpose may then mutually reinforce each other and spiral upward to enhance well-being. |
| Online social connectivity | Well-being | Virtual resilience: Online social connectivity and access to well-being resources can build "virtual resilience" and enhance well-being during stressful events like pandemics. |
| Sense of shared purpose and belonging | Well-being | A sense of shared purpose and belonging within your social groups is necessary for freedom, choice and self-determination to enhance well-being. |

The evaluation comprises three main stages. In the first stage, the hypotheses are generated by all contributors, including steps taken to ensure fairness and relevance for comparative analysis. In the second stage, the hypotheses from the first stage are independently and blindly reviewed by experts who represent the human academic community. These experts are asked to provide hypothesis ratings using a specially designed questionnaire to ensure statistical validity. The third stage delves deeper by transforming each research idea into the semantic space of a bidirectional encoder representation from transformers (BERT) (Lee et al., 2023), allowing us to intricately analyze the intrinsic reasons behind the rating disparities among the groups. This semantic mapping not only pinpoints the nuanced differences, but also provides potential insights into the cognitive constructs of each hypothesis.

### Evaluation procedure

**Selection of the focus area for hypothesis generation.** Selecting an appropriate focus area for hypothesis generation is crucial to ensure a balanced and insightful comparison of the hypothesis generation capacities between various contributors. In this study, our goal is to gauge the quality of hypotheses derived from four distinct contributors, with measures in place to mitigate potential confounding variables that might skew the results among groups (Rubin, 2005). Our choice of domain is informed by two pivotal criteria: the intricacy and subtlety of the subject matter and familiarity with the domain. It is essential that our chosen domain boasts sufficient complexity to prompt meaningful hypothesis generation and offer a robust assessment of both AI and human contributors" depth of understanding and creativity. Furthermore, while human contributors should be well-acquainted with the domain, their expertise need not match the vast corpus knowledge of the AI.

In terms of overarching human pursuits such as the search for happiness, positive psychology distinguishes itself by avoiding narrowly defined, individual-centric challenges (Seligman and Csikszentmihalyi, 2000). This alignment with our selection criteria is epitomized by well-being, a salient concept within positive psychology, as shown in Table 3. Well-being, with its multidimensional essence that encompass emotional, psychological, and social facets, and its central stature in both research and practical applications of positive psychology (Diener et al., 2010; Fredrickson, 2001; Seligman and Csikszentmihalyi, 2000), becomes the linchpin of our evaluation. The growing importance of well-being in the current global context offers myriad novel avenues for hypothesis generation and theoretical advancement (Forgeard et al., 2011; Madill et al., 2022; Otu et al., 2020). Adding to our rationale, the Positive Psychology Research Center at Tsinghua University is a globally renowned hub for cutting-edge research in this domain. Leveraging this stature, we secured participation from specialized Ph.D. students, reinforcing positive psychology as the most fitting domain for our inquiry.

**Hypotheses comparison.** In our study, the generated psychological hypotheses were categorized into four distinct groups, consisting of two experimental groups and two control groups. The experimental groups encapsulate hypotheses generated by our algorithm, either through random selection or handpicking by experts from a pool of generated hypotheses. On the other hand, control groups comprise research ideas that were meticulously crafted by doctoral students with substantial academic expertise in the domains and hypotheses generated by representative LLMs. In the following, we elucidate the methodology and underlying rationale for each group:

*LLMCG algorithm output (Random-selected LLMCG).* Following the requirement of generating hypotheses centred on well-being, the LLMCG algorithm crafted 130 unique hypotheses. These hypotheses were derived by LLMCG's evaluation of the most likely causal relationships related to well-being that had not been previously documented in research literature datasets. From this refined pool, 30 research ideas were chosen at random for this experimental group. These hypotheses represent the algorithm's ability to identify causal relationships and formulate pertinent hypotheses.

*LLMCG expert-vetted hypotheses (Expert-selected LLMCG).* For this group, two seasoned psychological researchers, one male aged 47 and one female aged 46, in-depth expertise in the realm of Positive Psychology, conscientiously handpicked 30 of the most promising hypotheses from the refined pool, excluding those from the Random-selected LLMCG category. The selection criteria centered on a holistic understanding of both the novelty and practical relevance of each hypothesis. With an illustrious post-doctoral journey and a robust portfolio of publications in positive psychology to their names, they rigorously sifted through the hypotheses, pinpointing those that showcased a perfect confluence of originality and actionable insight. These hypotheses were meticulously appraised for their relevance, structural coherence, and potential academic value, representing the nexus of machine intelligence and seasoned human discernment.

*PhD students' output (Control-Human).* We enlisted the expertise of 16 doctoral students from the Positive Psychology Research Center at Tsinghua University. Under the guidance of their supervisor, each student was provided with a questionnaire geared toward research on well-being. The participants were given a period of four working days to complete and return the questionnaire, which was distributed during vacation to ensure minimal external disruptions and commitments. The specific instructions provided in the questionnaire is detailed in Table B1, and each participant was asked to complete 3–4 research hypotheses. By the stipulated deadline, we received responses from 13 doctoral students, with a mean age of 31.92 years ($SD = 7.75$ years), cumulatively presenting 41 hypotheses related to well-being. To maintain uniformity with the other groups, a random selection was made to shortlist 30 hypotheses for further analysis. These hypotheses reflect the integration of core theoretical concepts with the latest insights into the domain, presenting an academic interpretation rooted in their rigorous training and education. Including this group in our study not only provides a natural benchmark for human ingenuity and expertise but also underscores the invaluable contribution of human cognition in research ideation, serving as a pivotal contrast to AI-generated hypotheses. This juxtaposition illuminates the nuanced differences between human intellectual depth and AI's analytical progress, enriching the comparative dimensions of our study.

*Claude model output (Control-Claude).* This group exemplifies the pinnacle of current LLM technology in generating research hypotheses. Since LLMCG is a nascent technology, its assessment requires a comparative study with well-established counterparts, creating a key paradigm in comparative research. Currently, Claude-2 and GPT-4 represent the apex of AI technology. For example, Claude-2, with an accuracy rate of 54. 4% excels in reasoning and answering questions, substantially outperforming other models such as Falcon, Koala and Vicuna, which have accuracy rates of 17.1–25.5% (Wu et al., 2023). To facilitate a more comprehensive evaluation of the new model by researchers and to increase the diversity and breadth of comparison, we chose Claude-2 as the control model. Using the detailed instructions provided in Table B2, Claude-2 was iteratively prompted to generate research hypotheses, generating ten hypotheses per prompt, culminating in a total of 50 hypotheses. Although the sheer number and range of these hypotheses accentuate the capabilities of Claude-2, to ensure compatibility in terms of complexity and depth between all groups, a subsequent refinement was considered essential. With minimal human intervention, GPT-4 was used to evaluate these 50 hypotheses and select the top 30 that exhibited the most innovative, relevant, and academically valuable insights. This process ensured the infusion of both the LLM's analytical prowess and a layer of qualitative rigor, thus giving rise to a set of hypotheses that not only align with the overarching theme of well-being but also resonate with current academic discourse.

**Hypotheses assessment**. The assessment of the hypotheses encompasses two key components: the evaluation conducted by eminent psychology professors emphasizing novelty and utility, and the deep semantic analysis involving BERT and $t$-distributed stochastic neighbor embedding ($t$-SNE) visualization to discern semantic structures and disparities among hypotheses.

*Human academic community*. The review task was entrusted to three eminent psychology professors (all male, mean age = 42.33), who have a decade-long legacy in guiding doctoral and master"s students in positive psychology and editorial stints in renowned journals; their task was to conduct a meticulous evaluation of the 120 hypotheses. Importantly, to ensure unbiased evaluation, the hypotheses were presented to them in a completely randomized order in the questionnaire.

Our emphasis was undeniably anchored to two primary tenets: novelty and utility (Cohen, 2017; Shardlow et al., 2018; Thompson and Skau, 2023; Yu et al., 2016), as shown in Table B3. Utility in hypothesis crafting demands that our propositions extend beyond mere factual accuracy; they must resonate deeply with academic investigations, ensuring substantial practical implications. Given the inherent challenges of research, marked by constraints in time, manpower, and funding, it is essential to design hypotheses that optimize the utilization of these resources. On the novelty front, we strive to introduce innovative perspectives that have the power to challenge and expand upon existing academic theories. This not only propels the discipline forward but also ensures that we do not inadvertently tread on ground already covered by our contemporaries.

*Deep semantic analysis*. While human evaluations provide invaluable insight into the novelty and utility of hypotheses, to objectively discern and visualize semantic structures and the disparities among them, we turn to the realm of deep learning. Specifically, we employ the power of BERT (Devlin et al., 2018). BERT, as highlighted by Lee et al. (2023), had a remarkable potential to assess the innovation of ideas. By translating each hypothesis into a high-dimensional vector in the BERT domain, we obtain the profound semantic core of each statement. However, such granularity in dimensions presents challenges when aiming for visualization.

To alleviate this and to intuitively understand the clustering and dispersion of these hypotheses in semantic space, we deploy the $t$-SNE ($t$-distributed Stochastic Neighbor Embedding) technique (Van der Maaten and Hinton, 2008), which is adept at reducing the dimensionality of the data while preserving the relative pairwise distances between the items. Thus, when we map our BERT-encoded hypotheses onto a 2D $t$-SNE plane, an immediate visual grasp on how closely or distantly related our hypotheses are in terms of their semantic content. Our intent is twofold: to understand the semantic terrains carved out by the different groups and to infer the potential reasons for some of the hypotheses garnered heightened novelty or utility ratings from experts. The convergence of human evaluations and semantic layouts, as delineated by Algorithm 1 in Appendix B, reveal the interplay between human intuition and the inherent semantic structure of the hypotheses.

**Results**

**Qualitative analysis by topic analysis**. To better understand the underlying thought processes and the topical emphasis of both PhD students and the LLMCG model, qualitative analyses were performed using visual tools such as word clouds and connection graphs, as detailed in Appendix B. The word cloud, as a graphical representation, effectively captures the frequency and importance of terms, providing direct visualization of the dominant themes. Connection graphs, on the other hand, elucidate the relationships and interplay between various themes and concepts. Using these visual tools, we aimed to achieve a more intuitive and clear representation of the data, allowing for easy comparison and interpretation.

Observations drawn from both the word clouds and the connection graphs in Figures B1 and B2 provide us with a rich tapestry of insights into the thought processes and priorities of Ph.D. students and the LLMCG model. For instance, the emphasis in the *Control-Human* word cloud on terms such as "robot" and "AI" indicates a strong interest among Ph.D. students in the nexus between technology and psychology. It is particularly fascinating to see a group of academically trained individuals focusing on the real world implications and intersections of their studies, as shown by their apparent draw toward trending topics. This not only underscores their adaptability but also emphasizes the importance of contextual relevance. Conversely, the LLMCG groups, particularly the *Expert-selected LLMCG* group, emphasize the community, collective experiences, and the nuances of social interconnectedness. This denotes a deep-rooted understanding and application of higher-order social psychological concepts, reflecting the model"s ability to dive deep into the intricate layers of human social behavior.

Furthermore, the connection graphs support these observations. The *Control-Human* graph, with its exploration of themes such as "Robot Companionship" and its relation to factors such as "heart rate variability (HRV)", demonstrates a confluence of technology and human well-being. The other groups, especially the *Random-selected LLMCG* group, yield themes that are more societal and structural, hinting at broader determinants of individual well-being.

**Analysis of human evaluations**. To quantify the agreement among the raters, we employed Spearman correlation coefficients. The results, as shown in Table B5, reveal a spectrum of agreement levels between the reviewer pairs, showcasing the subjective dimension intrinsic to the evaluation of novelty and usefulness. In particular, the correlation between reviewer 1 and reviewer 2 in novelty (Spearman $r = 0.387$, $p < 0.0001$) and between reviewer 2 and reviewer 3 in usefulness (Spearman $r = 0.376$, $p < 0.0001$) suggests a meaningful level of consensus, particularly highlighting their capacity to identify valuable insights when evaluating hypotheses.

The variations in correlation values, such as between reviewer 2 and reviewer 3 ($r = 0.069$, $p = 0.453$), can be attributed to the diverse research orientations and backgrounds of each reviewer. Reviewer 1 focuses on social ecology, reviewer 3 specializes in neuroscientific methodologies, and reviewer 2 integrates various views using technologies like virtual reality, and computational methods. In our evaluation, we present specific hypotheses cases to illustrate the differing perspectives between reviewers, as detailed in Table B4 and Figure B3. For example, C5 introduces the novel concept of "Virtual Resilience". Reviewers 1 and 3 highlighted its originality and utility, while reviewer 2 rated it lower in both categories. Meanwhile, C6, which focuses on social neuroscience, resonated with reviewer 3, while reviewers 1 and 2 only partially affirmed it. These differences underscore the complexity of evaluating scientific contributions and highlight the importance of considering a range of expert opinions for a comprehensive evaluation.

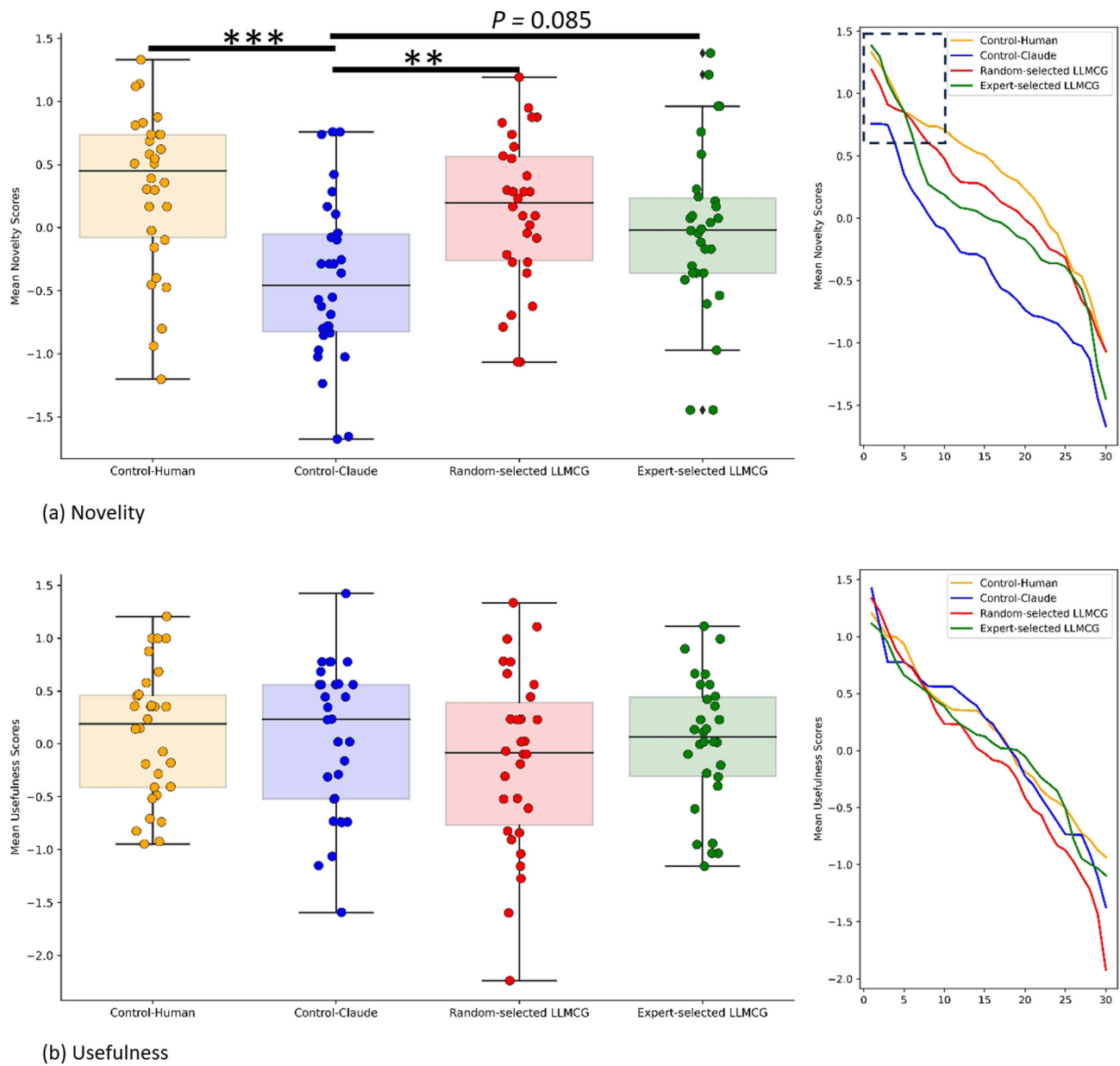

(a) Novelty

(b) Usefulness

**Fig. 2 Comparative analysis between groups.** Box plots on the left (**a**) and (**b**) depict distributions of novelty and usefulness scores, respectively, while smoothed line plots on the right demonstrate the descending order of novelty and usefulness scores and subjected to a moving average with a window size of 2. * denotes $p < 0.05$, ** denotes $p < 0.01$.

This assessment is divided into two main sections: Novelty analysis and usefulness analysis.

*Novelty analysis.* In the dynamic realm of scientific research, measuring and analyzing novelty is gaining paramount importance (Shin et al., 2022). ANOVA was used to analyze the novelty scores represented in Fig. 2a, and we identified a significant influence of the group factor on the mean novelty score between different reviewers. Initially, z-scores were calculated for each reviewer's ratings to standardize the scoring scale, which were then averaged. The distinct differences between the groups, as visualized in the boxplots, are statistically underpinned by the results in Table 5. The ANOVA results revealed a pronounced effect of the grouping factor ($F(3116) = 6.92$, $p = 0.0002$), with variance explained by the grouping factor (R-squared) of 15.19%.

Further pairwise comparisons using the Bonferroni method, as delineated in Table 5 and visually corroborated by Fig. 2a; significant disparities were discerned between *Random-selected LLMCG* and *Control-Claude* ($t(59) = 3.34, p = 0.007$) and between *Control-Human* and *Control-Claude* ($t(59) = 4.32, p < 0.001$). The Cohen's *d* values of 0.8809 and 1.1192 respectively indicate that the novelty scores for the *Random-selected LLMCG* and *Control-Human* groups are significantly higher than those for the *Control-Claude* group. Additionally, when considering the cumulative distribution plots to the right of Fig. 2a, we observe the distributional characteristics of the novel scores. For example, it can be observed that the *Expert-selected LLMCG* curve portrays a greater concentration in the middle score range when compared to the *Control-Claude*, curve but dominates in the high novelty scores (highlighted in dashed rectangle). Moreover, comparisons involving *Control-Human* with both *Random-selected LLMCG* and *Expert-selected LLMCG* did not manifest statistically significant variances, indicating aligned novelty perceptions among these groups. Finally, the comparisons between *Expert-selected LLMCG* and *Control-Claude* ($t(59) = 2.49$, $p = 0.085$) suggest a trend toward significance, with a Cohen's $d$ value of 0.6226 indicating generally higher novelty scores for *Expert-selected LLMCG* compared to *Control-Claude*.

Table 5 Bonferroni post-hoc tests for pairwise comparisons of novelty scores across different groups.

| Comparison | Contrast | Cohen's d | t value | p value |
|---|---|---|---|---|
| Mean Value | | | | |
| Control-Claude vs. Control-Human | 0.7148 | 1.1192 | 4.36 | <0.001 |
| Control-Claude vs. Random-selected LLMCG | 0.5478 | 0.8809 | 3.34 | 0.007 |
| Control-Claude vs. Expert-selected LLMCG | 0.4088 | 0.6226 | 2.49 | 0.085 |
| Median Value | | | | |
| Control-Claude vs. Control-Human | 0.7688 | 1.1031 | 4.01 | <0.001 |
| Control-Claude vs. Random-selected LLMCG | 0.6515 | 0.8875 | 3.40 | 0.006 |
| Control-Claude vs. Expert-selected LLMCG | 0.3264 | 0.4057 | 1.70 | 0.550 |
| Max Value | | | | |
| Control-Claude vs. Control-Human | 0.7815 | 1.1637 | 4.36 | <0.001 |
| Control-Claude vs. Random-selected LLMCG | 0.6231 | 1.0463 | 3.47 | 0.004 |
| Control-Claude vs. Expert-selected LLMCG | 0.5596 | 0.6987 | 3.12 | 0.014 |

To mitigate potential biases due to individual reviewer inclinations, we expanded our evaluation to include both median and maximum z-scores from the three reviewers for each hypothesis. These multifaceted analyses enhance the robustness of our results by minimizing the influence of extreme values and potential outliers. First, when analyzing the median novelty scores, the ANOVA test demonstrated a notable association with the grouping factor ($F(3,116) = 6.54$, $p = 0.0004$), which explained 14.41% of the variance. As illustrated in Table 5, pairwise evaluations revealed significant disparities between *Control-Human* and *Control-Claude* ($t(59) = 4.01$, $p = 0.001$), with *Control-Human* performing significantly higher than *Control-Claude* (Cohen's $d = 1.1031$). Similarly, there were significant differences between *Random-selected LLMCG* and *Control-Claude* ($t(59) = 3.40$, $p = 0.006$), where *Random-selected LLMCG* also significantly outperformed *Control-Claude* (Cohen's $d = 0.8875$). Interestingly, the comparison of *Expert-selected LLMCG* with *Control-Claude* ($t(59) = 1.70$, $p = 0.550$) and other group pairings did not include statistically significant differences.

Subsequently, turning our attention to maximum novelty scores provided crucial insights, especially where outlier scores may carry significant weight. The influence of the grouping factor was evident ($F(3,116) = 7.20$, $p = 0.0002$), indicating an explained variance of 15.70%. In particular, clear differences emerged between *Control-Human* and *Control-Claude* ($t(59) = 4.36$, $p < 0.001$), and between *Random-selected LLMCG* and *Control-Claude* ($t(59) = 3.47$, $p = 0.004$). A particularly intriguing observation was the significant difference between *Expert-selected LLMCG* and *Control-Claude* ($t(59) = 3.12$, $p = 0.014$). The Cohen's $d$ values of 1.1637, 1.0457, and 0.6987 respectively indicate that the novelty scores for the *Control-Human*, *Random-selected LLMCG*, and *Expert-selected LLMCG* groups are significantly higher than those for the *Control-Claude* group. Together, these analyses offer a multifaceted perspective on novelty evaluations. Specifically, the results of the median analysis echo and support those of the mean, reinforcing the reliability of our assessments. The discerned significance between *Control-Claude* and *Expert-selected LLMCG* in the median data emphasizes the intricate differences, while also pointing to broader congruence in novelty perceptions.

Table 6 Comparison of novelty and usefulness scores between the GPT-4 and LLMCG groups.

| Variable | Contrast | Cohen's d | t value | p Value |
|---|---|---|---|---|
| Novelty | | | | |
| Mean Value | 0.7311 | 1.2055 | 6.60 | <0.0001 |
| Median Value | 0.7954 | 1.1434 | 6.26 | <0.0001 |
| Max Value | 0.7501 | 1.1395 | 6.24 | <0.0001 |
| Usefulness | | | | |
| Mean Value | 0.1490 | 0.2387 | 1.31 | 0.1937 |
| Median Value | 0.1619 | 0.2045 | 1.12 | 0.2649 |
| Max Value | 0.0443 | 0.0793 | 0.43 | 0.6648 |

*Usefulness analysis*. Evaluating the practical impact of hypotheses is crucial in scientific research assessments. In the mean useful spectrum, the grouping factor did not exert a significant influence ($F(3,116) = 5.25$, $p = 0.553$). Figure 2b presents the utility score distributions between groups. The narrow interquartile range of *Control-Human* suggests a relatively consistent assessment among reviewers. On the other hand, the spread and outliers in the *Control-Claude* distribution hint at varied utility perceptions. Both LLMCG groups cover a broad score range, demonstrating a mixture of high and low utility scores, while the *Expert-selected LLMCG* gravitates more toward higher usefulness scores. The smoothed line plots accompanying Fig. 2b further detail the score densities. For instance, *Random-selected LLMCG* boasts several high utility scores, counterbalanced by a smattering of low scores. Interestingly, the distributions for *Control-Human* and *Expert-selected LLMCG* appear to be closely aligned. While mean utility scores provide an overarching view, the nuances within the boxplots and smoothed plots offer deeper insights. This comprehensive understanding can guide future endeavors in content generation and evaluation, spotlighting key areas of focus and potential improvements.

**Comparison between the LLMCG and GPT-4**. To evaluate the impact of integrating a causal graph with GPT-4, we performed an ablation study comparing the hypotheses generated by GPT-4 alone and those of the proposed LLMCG framework. For this experiment, 60 hypotheses were created using GPT-4, following the detailed instructions in Table B2. Furthermore, 60 hypotheses for the LLMCG group were randomly selected from the remaining pool of 70 hypotheses. Subsequently, both sets of hypotheses were assessed by three independent reviewers for novelty and usefulness, as previously described.

Table 6 shows a comparison between the GPT-4 and LLMCG groups, highlighting a significant difference in novelty scores

(mean value: $t(119) = 6.60$, $p < 0.0001$) but not in usefulness scores (mean value: $t(119) = 1.31$, $p = 0.1937$). This indicates that the LLMCG framework significantly enhances hypothesis novelty (all Cohen's $d > 1.1$) without affecting usefulness compared to the GPT-4 group. Figure B6 visually contrasts these findings, underlining the causal graph's unique role in fostering novel hypothesis generation when integrated with GPT-4.

**Deep semantic analysis**. The *t*-SNE visualizations (Fig. 3) illustrate the semantic relationships between different groups, capturing the patterns of novelty and usefulness. Notably, a distinct clustering among PhD students suggests shared academic influences, while the LLMCG groups display broader topic dispersion, hinting at a wider semantic understanding. The size of the bubbles reflects the novelty and usefulness scores, emphasizing the diverse perceptions of what is considered innovative versus beneficial. Additionally, the numbers near the yellow dots represent the participant IDs, which demonstrated that the semantics of the same participant, such as H05 or H06, are closely aligned. In Fig. B4, a distinct clustering of examples is observed, particularly highlighting the close proximity of hypotheses C3, C4, and C8 within the semantic space. This observation is further elucidated in Appendix B, enhancing the comprehension of BERT's semantic representation. Instead of solely depending on superficial textual descriptions, this analysis penetrates into the underlying understanding of concepts within the semantic space, a topic also explored in recent research (Johnson et al., 2023).

In the distribution of semantic distances (Fig. 4), we observed that the Control-Human group exhibits a distinctively greater semantic distance in comparison to the other groups, emphasizing their unique semantic orientations. The statistical support for this observation is derived from the ANOVA results, with a significant F-statistic ($F(3,1652) = 84.1611$, $p < 0.00001$), underscoring the impact of the grouping factor. This factor explains a remarkable 86.96% of the variance, as indicated by the *R*-squared value. Multiple comparisons, as shown in Table 7, further elucidate the subtleties of these group differences. *Control-Human* and *Control-Claude* exhibit a significant contrast in their semantic distances, as highlighted by the *t* value of 16.41 and the adjusted *p* value ($< 0.0001$). This difference indicates distinct thought patterns or emphasis in the two groups. Notably, *Control-Human* demonstrates a greater semantic distance (Cohen's $d = 1.1630$). Similarly, a comparison of the *Control-Claude* and LLMCG models reveals pronounced differences (Cohen's $d > 0.9$), more so with the *Expert-selected LLMCG* ($p < 0.0001$). A comparison of Control-Human with the LLMCG models shows divergent semantic orientations, with statistically significant larger distances than *Random-selected LLMCG* ($p = 0.0036$) and a trend toward difference with *Expert-selected LLMCG* ($p = 0.0687$). Intriguingly, the two LLMCG groups—Random-selected and Expert-selected—exhibit similar semantic distances, as evidenced by a nonsignificant $p$ value of 0.4362. Furthermore, the significant distinctions we observed, particularly between the *Control-Human* and other groups, align with human evaluations of novelty. This coherence indicates that the BERT space representation coupled with statistical analyses could effectively mimic human judgment. Such results underscore the potential of this approach for automated hypothesis testing, paving the way for more efficient and streamlined semantic evaluations in the future.

In general, visual and statistical analyses reveal the nuanced semantic landscapes of each group. While the Ph.D. students' shared background influences their clustering, the machine models exhibit a comprehensive grasp of topics, emphasizing the intricate interplay of individual experiences, academic influences, and algorithmic understanding in shaping semantic representations.

This investigation carried out a detailed evaluation of the various hypothesis contributors, blending both quantitative and qualitative analyses. In terms of topic analysis, distinct variations were observed between *Control-Human* and LLMCG, the latter presenting more expansive thematic coverage. For human evaluation, hypotheses from Ph.D. students paralleled the LLMCG in novelty, reinforcing AI's growing competence in mirroring human innovative thinking. Furthermore, when juxtaposed with AI models such as *Control-Claude*, the LLMCG exhibited increased novelty. Deep semantic analysis via *t*-SNE and BERT representations allowed us to intuitively grasp semantic essence of hypotheses, signaling the possibility of future automated hypothesis assessments. Interestingly, LLMCG appeared to encompass broader complementary domains compared to human input. Taken together, these findings highlight the emerging role of AI in hypothesis generation and provide key insights into hypothesis evaluation across diverse origins.

**General discussion**

This research delves into the synergistic relationship between LLM and causal graphs in the hypothesis generation process. Our findings underscore the ability of LLM, when integrated with causal graph techniques, to produce meaningful hypotheses with increased efficiency and quality. By centering our investigation on "well-being" we emphasize its pivotal role in psychological studies and highlight the potential convergence of technology and society. A multifaceted assessment approach to evaluate quality by topic analysis, human evaluation and deep semantic analysis demonstrates that AI-augmented methods not only outshine LLM-only techniques in generating hypotheses with superior novelty and show quality on par with human expertise but also boast the capability for more profound conceptual incorporations and a broader semantic spectrum. Such a multifaceted lens of assessment introduces a novel perspective for the scholarly realm, equipping researchers with an enriched understanding and an innovative toolset for hypothesis generation. At its core, the melding of LLM and causal graphs signals a promising frontier, especially in regard to dissecting cornerstone psychological constructs such as "well-being". This marriage of methodologies, enriched by the comprehensive assessment angle, deepens our comprehension of both the immediate and broader ramifications of our research endeavors.

The prominence of causal graphs in psychology is profound, they offer researchers a unified platform for synthesizing and hypothesizing diverse psychological realms (Borsboom et al., 2021; Uleman et al., 2021). Our study echoes this, producing groundbreaking hypotheses comparable in depth to early expert propositions. Deep semantic analysis bolstered these findings, emphasizing that our hypotheses have distinct cross-disciplinary merits, particularly when compared to those of individual doctoral scholars. However, the traditional use of causal graphs in psychology presents challenges due to its demanding nature, often requiring insights from multiple experts (Crielaard et al., 2022). Our research, however, harnesses LLM's causal extraction, automating causal pair derivation and, in turn, minimizing the need for extensive expert input. The union of the causal graphs' systematic approach with AI-driven creativity, as seen with LLMs, paves the way for the future of psychological inquiry. Thanks to advancements in AI, barriers once created by causal graphs' intricate procedures are being dismantled. Furthermore, as the era of big data dawns, the integration of AI and causal graphs in psychology augments research capabilities, but also brings into focus the broader implications for society. This fusion provides a nuanced understanding of the intricate sociopsychological dynamics, emphasizing the importance of adapting research methodologies in tandem with technological progress.

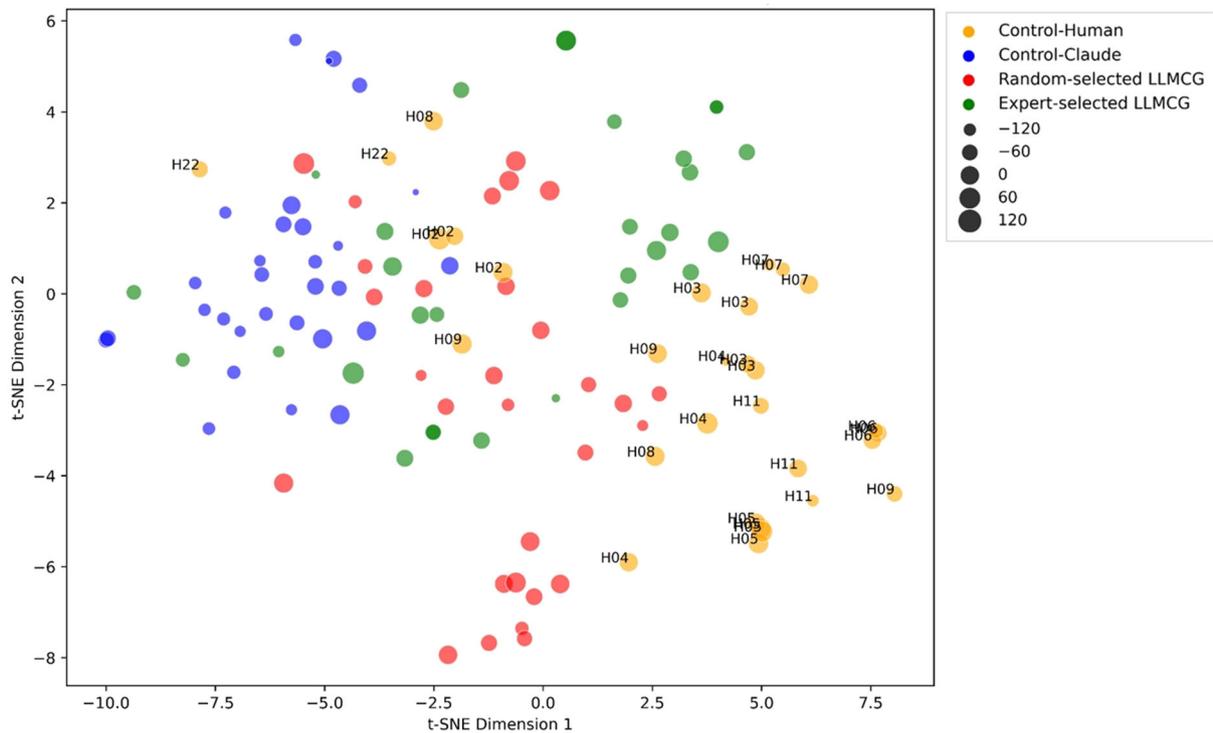

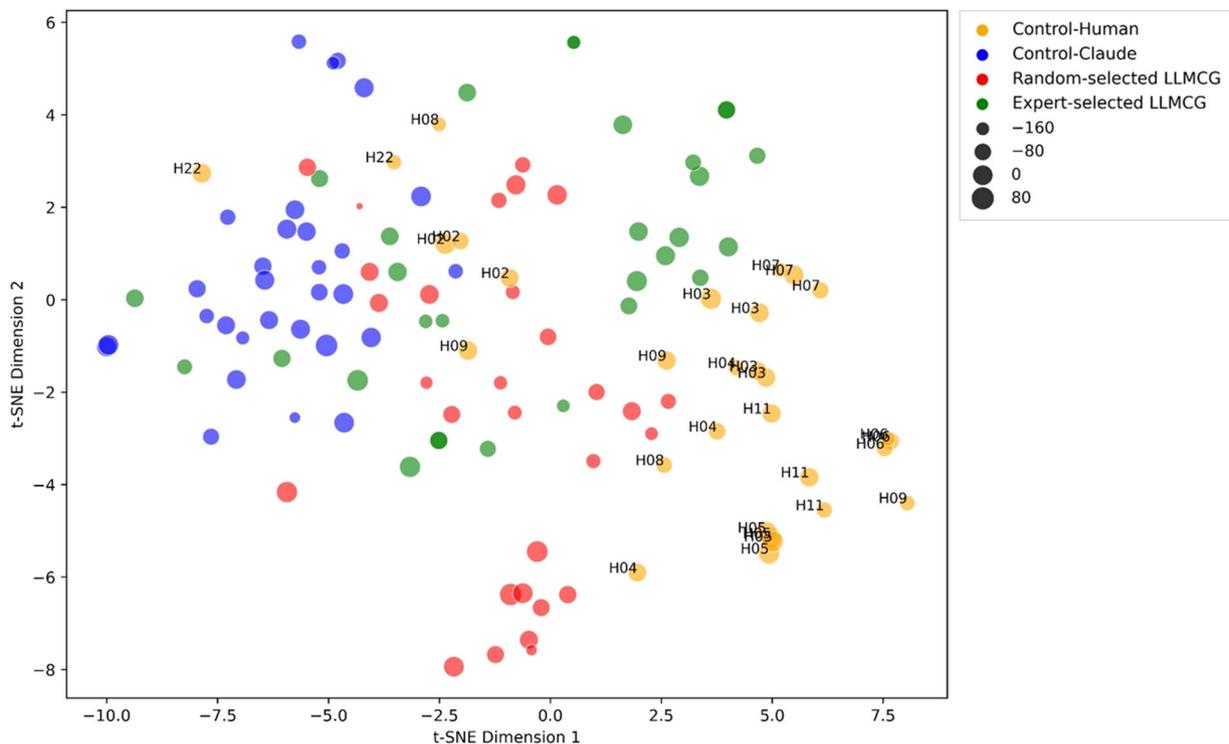

**Fig. 3 The *t*-SNE visualization of semantic representations.** Comparison of (**a**) novelty and (**b**) usefulness scores (bubble size scaled by 100) among the different groups.

In the realm of research, LLMs serve a unique purpose, often by acting as the foundation or baseline against which newer methods and approaches are assessed. The demonstrated productivity enhancements by generative AI tools, as evidenced by

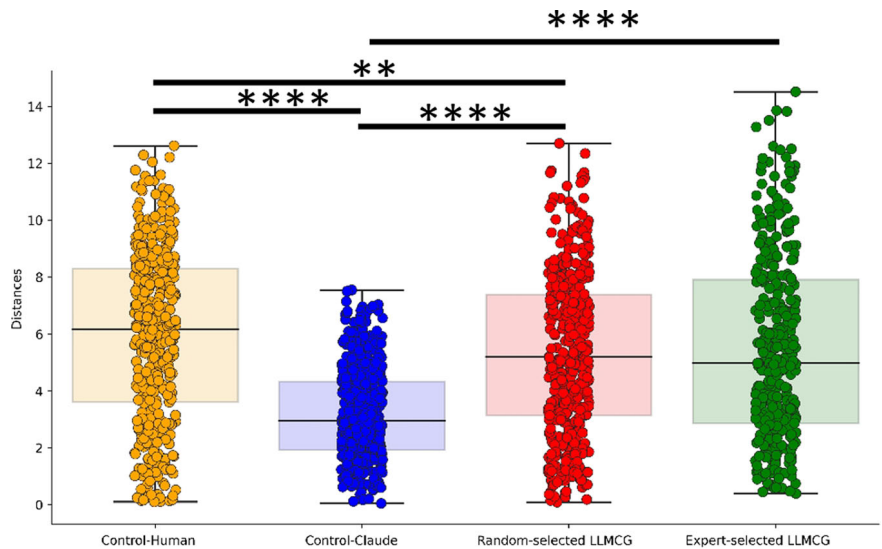

**Fig. 4 Distribution of semantic distances among different groups in BERT space.** Note: ** denotes $p < 0.01$, **** denotes $p < 0.0001$.

Table 7 Bonferroni post-hoc tests for pairwise comparisons of deep semantic distance across different groups.

| Comparison | Contrast | Cohen's d | t value | p value |
|---|---|---|---|---|
| Control-Claude vs. Control-Human | 2.6949 | 1.1630 | 16.41 | <0.0001 |
| Control-Claude vs. Random-selected LLMCG | 2.1175 | 1.0701 | 13.45 | <0.0001 |
| Control-Claude vs. Expert-selected LLMCG | 2.2870 | 0.9150 | 12.72 | <0.0001 |
| Random-selected LLMCG vs. Control-Human | 0.5773 | 0.1396 | 2.91 | 0.0036 |
| Expert-selected LLMCG vs. Control-Human | 0.4078 | 0.1094 | 1.82 | 0.0687 |
| Random-selected LLMCG vs. Expert-selected LLMCG | 0.1695 | 0.0152 | 0.78 | 0.4362 |

Noy and Zhang (2023), indicate the potential of such LLMs. In our investigation, we pit the hypotheses generated by such substantial models against our integrated LLMCG approach. Intriguingly, while these LLMs showcased admirable practicality in their hypotheses, they substantially lagged behind in terms of innovation when juxtaposed with the doctoral student and LLMCG group. This divergence in results can be attributed to the causal network curated from 43k research papers, funneling the vast knowledge reservoir of the LLM squarely into the realm of scientific psychology. The increased precision in hypothesis generation by these models fits well within the framework of generative networks. Tong et al. (2021) highlighted that, by integrating structured constraints, conventional neural networks can accurately generate semantically relevant content. One of the salient merits of the causal graph, in this context, is its ability to alleviate inherent ambiguity or interpretability challenges posed by LLMs. By providing a systematic and structured framework, the causal graph aids in unearthing the underlying logic and rationale of the outputs generated by LLMs. Notably, this finding echoes the perspective of Pan et al. (2023), where the integration of structured knowledge from knowledge graphs was shown to provide an invaluable layer of clarity and interpretability to LLMs, especially in complex reasoning tasks. Such structured approaches not only boost the confidence of researchers in the hypotheses derived but also augment the transparency and understandability of LLM outputs. In essence, leveraging causal graphs may very well herald a new era in model interpretability, serving as a conduit to unlock the black box that large models often represent in contemporary research.

In the ever-evolving tapestry of research, every advancement invariably comes with its unique set of constraints, and our study was no exception. On the technical front, a pivotal challenge stemmed from the opaque inner workings of the GPT. Determining the exact machinations within the GPT that lead to the formation of specific causal pairs remains elusive, thereby reintroducing the age-old issue of AI's inherent lack of transparency (Buruk, 2023; Cao and Yousefzadeh, 2023). This opacity is magnified in our sparse causal graph, which, while expansive, is occasionally riddled with concepts that, while semantically distinct, converge in meaning. In tangible applications, a careful and meticulous algorithmic evaluation would be imperative to construct an accurate psychological conceptual landscape. Delving into psychology, which bridges humanities and natural sciences, it continuously aims to unravel human cognition and behavior (Hergenhahn and Henley, 2013). Despite the dominance of traditional methodologies (Henrich et al., 2010; Shah et al., 2015), the present data-centric era amplifies the synergy of technology and humanities, resonating with Hasok Chang's vision of enriched science (Chang, 2007). This symbiosis is evident when assessing structural holes in social networks (Burt, 2004) and viewing novelty as a bridge across these divides (Foster et al., 2021). Such perspectives emphasize the importance of thorough algorithmic assessments, highlighting potential avenues in humanities research, especially when incorporating large language models for innovative hypothesis crafting and verification.

However, there are some limitations to this research. Firstly, we acknowledge that constructing causal relationship graphs has potential inaccuracies, with ~13% relationship pairs not aligning with human expert estimations. Enhancing the estimation of relationship extraction could be a pathway to improve the accuracy of the causal graph, potentially leading to more robust hypotheses. Secondly, our validation process was limited to 130 hypotheses, however, the vastness of our conceptual landscape suggests countless possibilities. As an exemplar, the twenty pivotal psychological concepts highlighted in Table 3 alone could spawn an extensive array of hypotheses. However, the validation of these surrounding hypotheses would unquestionably lead to a multitude of speculations. A striking observation during our validation was the inconsistency in the evaluations of the senior expert panels (as shown in Table B5). This shift underscores a pivotal insight: our integration of AI has transitioned the dependency on scarce expert resources from hypothesis generation to evaluation. In the future, rigorous evaluations ensuring both novelty and utility could become a focal point of exploration. The promising path forward necessitates a thoughtful integration of technological innovation and human expertise to fully realize the potential suggested by our study.

In conclusion, our research provides pioneering insight into the symbiotic fusion of LLMs, which are epitomized by GPT, and causal graphs from the realm of psychological hypothesis generation, especially emphasizing "well-being". Importantly, as highlighted by (Cao and Yousefzadeh, 2023), ensuring a synergistic alignment between domain knowledge and AI extrapolation is crucial. This synergy serves as the foundation for maintaining AI models within their conceptual limits, thus bolstering the validity and reliability of the hypotheses generated. Our approach intricately interweaves the advanced capabilities of LLMs with the methodological prowess of causal graphs, thereby optimizing while also refining the depth and precision of hypothesis generation. The causal graph, of paramount importance in psychology due to its cross-disciplinary potential, often demands vast amounts of expert involvement. Our innovative approach addresses this by utilizing LLM's exceptional causal extraction abilities, effectively facilitating the transition of intense expert engagement from hypothesis creation to evaluation. Therefore, our methodology combined LLM with causal graphs, propelling psychological research forward by improving hypothesis generation and offering tools to blend theoretical and data-centric approaches. This synergy particularly enriches our understanding of social psychology's complex dynamics, such as happiness research, demonstrating the profound impact of integrating AI with traditional research frameworks.

### Data availability

### References

Battleday RM, Peterson JC, Griffiths TL (2020) Capturing human categorization of natural images by combining deep networks and cognitive models. Nat Commun 11(1):5418
Bechmann A, Bowker GC (2019) Unsupervised by any other name: hidden layers of knowledge production in artificial intelligence on social media. Big Data Soc 6(1):2053951718819569
Binz M, Schulz E (2023) Using cognitive psychology to understand GPT-3. Proc Natl Acad Sci 120(6):e2218523120
Boden MA (2009) Computer models of creativity. AI Mag 30(3):23–23
Borsboom D, Deserno MK, Rhemtulla M, Epskamp S, Fried EI, McNally RJ (2021) Network analysis of multivariate data in psychological science. Nat Rev Methods Prim 1(1):58
Burt RS (2004) Structural holes and good ideas. Am J Sociol 110(2):349–399
Buruk O (2023) Academic writing with GPT-3.5: reflections on practices, efficacy and transparency. arXiv preprint arXiv:2304.11079
Cao X, Yousefzadeh R (2023) Extrapolation and AI transparency: why machine learning models should reveal when they make decisions beyond their training. Big Data Soc 10(1):20539517231169731
Chang H (2007) Scientific progress: beyond foundationalism and coherentism1. R Inst Philos Suppl 61:1–20
Cheng K, Guo Q, He Y, Lu Y, Gu S, Wu H (2023) Exploring the potential of GPT-4 in biomedical engineering: the dawn of a new era. Ann Biomed Eng 51:1645–1653
Cichy RM, Khosla A, Pantazis D, Torralba A, Oliva A (2016) Comparison of deep neural networks to spatio-temporal cortical dynamics of human visual object recognition reveals hierarchical correspondence. Sci Rep 6(1):27755
Cohen BA (2017) How should novelty be valued in science? Elife 6:e28699
Crielaard L, Uleman JF, Châtel BD, Epskamp S, Sloot P, Quax R (2022) Refining the causal loop diagram: a tutorial for maximizing the contribution of domain expertise in computational system dynamics modeling. Psychol Methods 29(1):169–201
Devlin J, Chang M W, Lee K & Toutanova (2019) Bert: pre-training of deep bidirectional transformers for language understanding. In Proceedings of the 2019 Conference of the North American Chapter of the Association for Computational Linguistics: Human Language Technologies, Volume 1 (Long and Short Papers) (pp. 4171–4186)
Diener E, Wirtz D, Tov W, Kim-Prieto C, Choi D-W, Oishi S, Biswas-Diener R (2010) New well-being measures: short scales to assess flourishing and positive and negative feelings. Soc Indic Res 97:143–156
Dowling M, Lucey B (2023) ChatGPT for (finance) research: the Bananarama conjecture. Financ Res Lett 53:103662
Forgeard MJ, Jayawickreme E, Kern ML, Seligman ME (2011) Doing the right thing: measuring wellbeing for public policy. Int J Wellbeing 1(1):79–106
Foster J G, Shi F & Evans J (2021) Surprise! Measuring novelty as expectation violation. SocArXiv
Fredrickson BL (2001) The role of positive emotions in positive psychology: The broaden-and-build theory of positive emotions. Am Psychol 56(3):218
Gu Q, Kuwajerwala A, Morin S, Jatavallabhula K M, Sen B, Agarwal, A et al. (2024) ConceptGraphs: open-vocabulary 3D scene graphs for perception and planning. In 2nd Workshop on Language and Robot Learning: Language as Grounding
Henrich J, Heine SJ, Norenzayan A (2010) Most people are not WEIRD. Nature 466(7302):29–29
Hergenhahn B R, Henley T (2013) An introduction to the history of psychology. Cengage Learning
Jaccard J, Jacoby J (2019) Theory construction and model-building skills: a practical guide for social scientists. Guilford publications
Johnson DR, Kaufman JC, Baker BS, Patterson JD, Barbot B, Green AE (2023) Divergent semantic integration (DSI): Extracting creativity from narratives with distributional semantic modeling. Behav Res Methods 55(7):3726–3759
Kıcıman E, Ness R, Sharma A & Tan C (2023) Causal reasoning and large language models: opening a new frontier for causality. arXiv preprint arXiv:2305.00050
Koehler DJ (1994) Hypothesis generation and confidence in judgment. J Exp Psychol Learn Mem Cogn 20(2):461–469
Krenn M, Zeilinger A (2020) Predicting research trends with semantic and neural networks with an application in quantum physics. Proc Natl Acad Sci 117(4):1910–1916
Lee H, Zhou W, Bai H, Meng W, Zeng T, Peng K & Kumada T (2023) Natural language processing algorithms for divergent thinking assessment. In: Proc IEEE 6th Eurasian Conference on Educational Innovation (ECEI) p 198–202
Madill A, Shloim N, Brown B, Hugh-Jones S, Plastow J, Setiyawati D (2022) Mainstreaming global mental health: Is there potential to embed psychosocial well-being impact in all global challenges research? Appl Psychol Health Well-Being 14(4):1291–1313
McCarthy M, Chen CC, McNamee RC (2018) Novelty and usefulness trade-off: cultural cognitive differences and creative idea evaluation. J Cross-Cult Psychol 49(2):171–198
McGuire WJ (1973) The yin and yang of progress in social psychology: seven koan. J Personal Soc Psychol 26(3):446–456
Miron-Spektor E, Beenen G (2015) Motivating creativity: The effects of sequential and simultaneous learning and performance achievement goals on product novelty and usefulness. Organ Behav Hum Decis Process 127:53–65
Nisbett RE, Peng K, Choi I, Norenzayan A (2001) Culture and systems of thought: holistic versus analytic cognition. Psychol Rev 108(2):291–310
Noy S, Zhang W (2023) Experimental evidence on the productivity effects of generative artificial intelligence. Science 381:187–192
Oleinik A (2019) What are neural networks not good at? On artificial creativity. Big Data Soc 6(1):2053951719839433
Otu A, Charles CH, Yaya S (2020) Mental health and psychosocial well-being during the COVID-19 pandemic: the invisible elephant in the room. Int J Ment Health Syst 14:1–5
Pan S, Luo L, Wang Y, Chen C, Wang J & Wu X (2024) Unifying large language models and knowledge graphs: a roadmap. IEEE Transactions on Knowledge and Data Engineering 36(7):3580–3599
Rubin DB (2005) Causal inference using potential outcomes: design, modeling, decisions. J Am Stat Assoc 100(469):322–331

Sanderson K (2023) GPT-4 is here: what scientists think. Nature 615(7954):773

Seligman ME, Csikszentmihalyi M (2000) Positive psychology: an introduction. Am Psychol 55(1):5–14

Shah DV, Cappella JN, Neuman WR (2015) Big data, digital media, and computational social science: possibilities and perils. Ann Am Acad Political Soc Sci 659(1):6–13

Shardlow M, Batista-Navarro R, Thompson P, Nawaz R, McNaught J, Ananiadou S (2018) Identification of research hypotheses and new knowledge from scientific literature. BMC Med Inform Decis Mak 18(1):1–13

Shin H, Kim K, Kogler DF (2022) Scientific collaboration, research funding, and novelty in scientific knowledge. PLoS ONE 17(7):e0271678

Thomas RP, Dougherty MR, Sprenger AM, Harbison J (2008) Diagnostic hypothesis generation and human judgment. Psychol Rev 115(1):155–185

Thomer AK, Wickett KM (2020) Relational data paradigms: what do we learn by taking the materiality of databases seriously? Big Data Soc 7(1):2053951720934838

Thompson WH, Skau S (2023) On the scope of scientific hypotheses. R Soc Open Sci 10(8):230607

Tong S, Liang X, Kumada T, Iwaki S (2021) Putative ratios of facial attractiveness in a deep neural network. Vis Res 178:86–99

Uleman JF, Melis RJ, Quax R, van der Zee EA, Thijssen D, Dresler M (2021) Mapping the multicausality of Alzheimer's disease through group model building. GeroScience 43:829–843

Van der Maaten L, Hinton G (2008) Visualizing data using t-SNE. J Mach Learn Res 9(11):2579–2605

Vaswani A, Shazeer N, Parmar N, Uszkoreit J, Jones L, Gomez A N & Polosukhin I (2017) Attention is all you need. In *Advances in Neural Information Processing Systems*

Wang H, Fu T, Du Y, Gao W, Huang K, Liu Z (2023) Scientific discovery in the age of artificial intelligence. Nature 620(7972):47–60

Webber J (2012) A programmatic introduction to neo4j. In *Proceedings of the 3rd annual conference on systems, programming, and applications: software for humanity* p 217–218

Williams K, Berman G, Michalska S (2023) Investigating hybridity in artificial intelligence research. Big Data Soc 10(2):20539517231180577

Wu S, Koo M, Blum L, Black A, Kao L, Scalzo F & Kurtz I (2023) A comparative study of open-source large language models, GPT-4 and Claude 2: multiple-choice test taking in nephrology. *arXiv preprint arXiv:2308.04709*

Yu F, Peng T, Peng K, Zheng SX, Liu Z (2016) The Semantic Network Model of creativity: analysis of online social media data. Creat Res J 28(3):268–274

## Acknowledgements

The authors thank Dr. Honghong Bai (Radboud University), Dr. ChienTe Wu (The University of Tokyo), Dr. Peng Cheng (Tsinghua University), and Yusong Guo (Tsinghua University) for their great comments on the earlier version of this manuscript. This research has been generously funded by personal contributions, with special acknowledgment to K. Mao. Additionally, he conceived and developed the causality graph and AI hypothesis generation technology presented in this paper from scratch, and generated all AI hypotheses and paid for its costs. The authors sincerely thank K. Mao for his support, which enabled this research. In addition, K. Peng and S. Tong were partly supported by the Tsinghua University lnitiative Scientific Research Program (No. 20213080008), Self-Funded Project of Institute for Global Industry, Tsinghua University (202-296-001), Shuimu Scholars program of Tsinghua University (No. 2021SM157), and the China Postdoctoral International Exchange Program (No. YJ20210266).

## Author contributions

Song Tong: Data analysis, Experiments, Writing—original draft & review. Kai Mao: Designed the causality graph methodology, Generated AI hypotheses, Developed hypothesis generation techniques, Writing—review & editing. Zhen Huang: Statistical Analysis, Experiments, Writing—review & editing. Yukun Zhao: Conceptualization, Project administration, Supervision, Writing—review & editing. Kaiping Peng: Conceptualization, Writing—review & editing.

## Competing interests

The author(s) declared no potential conflicts of interest with respect to the research, authorship, and/or publication of this article.

## Ethical approval

In this study, ethical approval was granted by the Institutional Review Board (IRB) of the Department of Psychology at Tsinghua University, China. The Research Ethics Committee documented this approval under the number IRB202306, following an extensive review that concluded on March 12, 2023. This approval indicates the research's strict compliance with the IRB's guidelines and regulations, ensuring ethical integrity and adherence throughout the study.

## Informed consent

Before participating, all study participants gave their informed consent. They received comprehensive details about the study's goals, methods, potential risks and benefits, confidentiality safeguards, and their rights as participants. This process guaranteed that participants were fully informed about the study's nature and voluntarily agreed to participate, free from coercion or undue influence.

## Additional information

**Supplementary information** The online version contains supplementary material available at https://doi.org/10.1057/s41599-024-03407-5.

**Correspondence** and requests for materials should be addressed to Yukun Zhao or Kaiping Peng.

**Reprints and permission information** is available at http://www.nature.com/reprints

**Publisher's note** Springer Nature remains neutral with regard to jurisdictional claims in published maps and institutional affiliations.